\documentclass[10pt,twocolumn,letterpaper]{article}

\usepackage[pagenumbers]{cvpr} 

\newcommand{\todo}[1]{{\color{red}#1}}

\definecolor{cvprblue}{rgb}{0.21,0.49,0.74}
\usepackage[pagebackref,breaklinks,colorlinks,allcolors=cvprblue]{hyperref}

\def\confName{CVPR}
\def\confYear{2027}

\usepackage{dsfont}
\usepackage{colortbl}  
\usepackage[skins]{tcolorbox}  

\title{RTime-QA: A Benchmark for Atomic Temporal Event Understanding in Multimodal Large Language Models}

\author{%
    Yuqi Liu$^{1}$ \hspace{1pt}
    Qin Jin$^{2}$ \hspace{1pt}
    Tianyuan Qu$^{1}$ \hspace{1pt}
    Xuan Liu$^{2}$ \hspace{1pt}
    Yang Du$^{2}$ \hspace{1pt}
    Bei Yu$^{1}$ \hspace{1pt}
    Jiaya Jia$^{3}$ \hspace{1pt}
    \\ 
    \small The Chinese University of Hong Kong$^{1}$ \hspace{3pt} Renmin University of China$^{2}$ \hspace{3pt}  The Hong Kong University of Science and Technology$^{3}$ \\
}

\begin{document}
\twocolumn[{%
\renewcommand\twocolumn[1][]{#1}%
\maketitle
\providecommand{\Vr}{\overline{V}}
\providecommand{\rtqa}{RTime-QA}
\providecommand{\rtqafull}{RTime-QA-Full}
\providecommand{\rtqatiny}{RTime-QA-Tiny}
\providecommand{\rtqacore}{RTime-QA-Core}
\providecommand{\rtqacomp}{RTime-QA-Composite}
\begin{center}
  \centering
  \includegraphics[width=\textwidth]{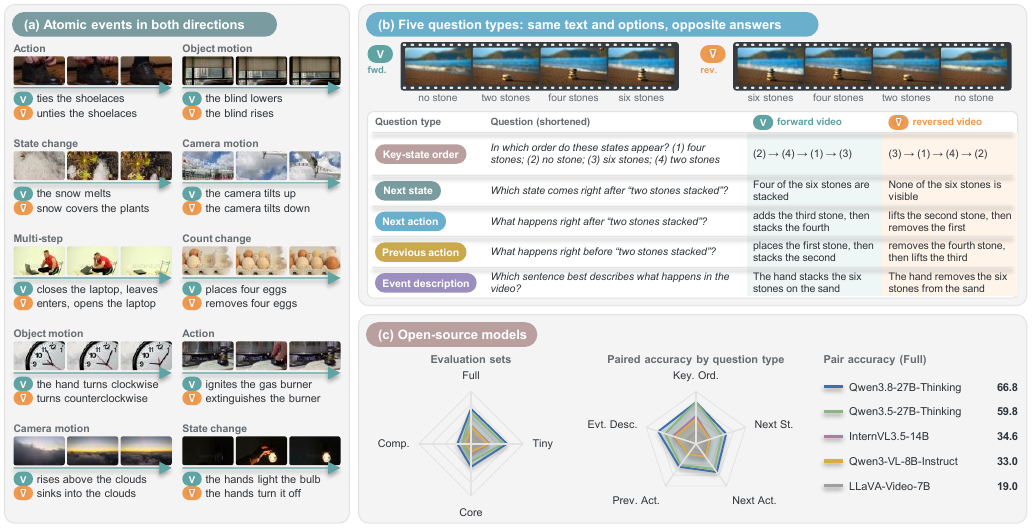}
  \captionof{figure}{\rtqa{} at a glance. (a)~Atomic events of the six event categories: key-state frames of each forward
  video $V$ and the events that $V$ and its reversal $\Vr$ depict. (b)~The five question types on one video pair
  (questions and answers shortened; each question has four options). Every question is asked on $V$ and on $\Vr$ with
  identical text and options, but the correct answers differ. (c)~Open-source models: paired accuracy on \rtqafull,
  \rtqatiny{} and \rtqacore{} and group accuracy on \rtqacomp{} (left), and paired accuracy per question type on
  \rtqafull{} (middle; Key.\ Ord.: key-state order; Next St.: next state; Next Act.: next action; Prev.\ Act.: previous
  action; Evt.\ Desc.: event description).}
  \label{fig:teaser}
\end{center}
}]

\providecommand{\Vr}{\overline{V}}
\providecommand{\qr}{\bar{q}}
\providecommand{\rtqa}{RTime-QA}
\providecommand{\rtqatiny}{RTime-QA-Tiny}
\providecommand{\rtqacore}{RTime-QA-Core}
\providecommand{\rtqafull}{RTime-QA-Full}
\providecommand{\rtqacomp}{RTime-QA-Composite}
\definecolor{rtqarank}{HTML}{7AB5B4}
\providecommand{\rankone}[1]{\cellcolor{rtqarank!70}#1}
\providecommand{\ranktwo}[1]{\cellcolor{rtqarank!40}#1}
\providecommand{\rankthree}[1]{\cellcolor{rtqarank!18}#1}
\providecommand{\rtqatrain}{RTime-QA-Train}
\providecommand{\rtqatrainQ}{8,966}
\providecommand{\rtqatrainV}{1,352}
\providecommand{\todo}[1]{{\color{red}#1}}

\begin{abstract}
Multimodal large language models (MLLMs) perform well on video benchmarks, yet many video questions can be answered from
a single frame or from language priors, which leaves open whether these models perceive how an event unfolds in
time. We introduce \rtqa, a benchmark for atomic temporal event understanding built on time-reversed question
pairs: each video is accompanied by its reversal, and each question is asked on both videos with identical text and
options but different correct answers, so that a deterministic model that ignores the order of the frames cannot answer both.
\rtqa{} comprises 8,420 questions of five types on 1,278 videos, 639 videos and their reversals. The questions are derived
from event annotations that a proposer model writes, a reviewer model checks and human annotators verify, and three
subsets target fast evaluation (\rtqatiny), questions that are hard to answer from the first and last frames (\rtqacore) and two
consecutive events (\rtqacomp). All 20 evaluated models are less accurate on reversed videos.
On \rtqafull, proprietary models answer both questions of 90.6--94.4\% of the pairs and the strongest open-source
model of 66.8\%; on \rtqacore, the proprietary models drop to 55.6--64.5\%, and shuffling the frames reduces the paired
accuracy of the tested open-source models to 3.4--7.4\%.
We further build \rtqatrain{} with the same pipeline, without human verification, and train three open-source MLLMs on it with supervised fine-tuning (SFT) and reinforcement learning (DAPO) under matched settings: both raise paired accuracy on \rtqafull, SFT by 20.7--27.4 points and DAPO by 5.9--22.9 points. For Qwen3.5-9B-Thinking, the training data also help on two other temporal benchmarks: SFT and DAPO raise the Vinoground text score by 5.8 and 3.8 points, respectively, and DAPO raises the TVBench macro accuracy by 4.6 points.
\end{abstract}

\section{Introduction}
\label{sec:intro}

Videos record events that unfold over time. Understanding an atomic event, such as a person tying a shoelace or
lowering a roller blind, requires perceiving its intermediate states in order, the steps between them and its
direction in time; played backwards, the same frames show the shoelace being untied and the
blind being raised. Multimodal large language models (MLLMs) score highly on comprehensive video
benchmarks~\cite{fu2024videomme,li2024mvbench,mangalam2023egoschema,xiao2021nextqa}, yet it is unclear whether
they perceive this temporal structure or rely on other cues.

Other cues often suffice. Many video-language tasks can be solved from a single
frame~\cite{buch2022revisiting,lei2022singleframe}, and the text of widely used video benchmarks often reveals the
answer without visual input~\cite{cores2024tvbench}. Temporal benchmarks therefore target order, direction and speed,
or contrast captions that differ only in a temporal
concept~\cite{liu2024tempcompass,li2024vitatecs,zhang2024vinoground,bagad2023testoftime,patraucean2023perceptiontest},
and MLLMs struggle with the direction of time~\cite{xue2025arrowoftime}. A question on a single video, however,
can still be answered from the usual direction of such events: without frames, three open-source models answer 42.2--51.5\% of our questions on forward videos but only
28.7--30.4\% on reversed videos (\cref{sec:exp_analysis}).

\begin{figure*}[t]
  \centering
  \includegraphics[width=\linewidth]{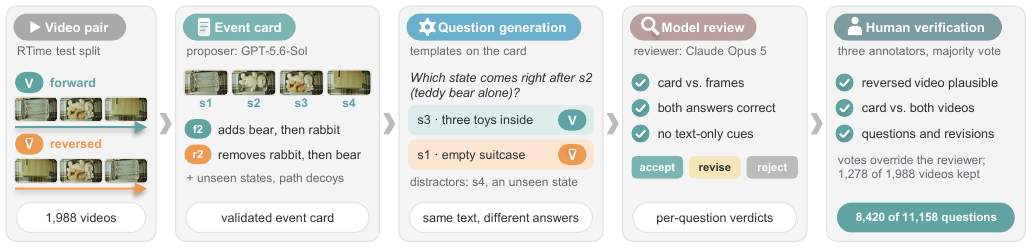}
  \caption{Construction pipeline of \rtqa, illustrated on one video pair; the last panel gives the overall filtering
  (\cref{sec:construction}).}
  \label{fig:pipeline}
\end{figure*}

To remove these shortcuts by design, we propose \rtqa.
Building on the reversible videos of RTime~\cite{rtime}, it pairs each video $V$ with its time-reversed counterpart
$\Vr$ and poses every question on both, where the same text and options have different correct answers
(\cref{fig:teaser}). Paired accuracy credits a pair only if both questions are answered correctly, so deterministic
shortcuts that ignore the frames or their order score zero. A proposer model annotates each
video with an \emph{event card} of four key states and the steps between them, templates derive questions of five
types from the card, a reviewer model checks every card and question against the frames, and human annotators verify
every video (\cref{fig:pipeline}). \rtqa{} contains 8,420 questions on 1,278 videos (639 videos and their reversals).
Besides this full set, \rtqafull, we provide \rtqatiny{} (2,036 questions) for fast evaluation, \rtqacore{}
(428 questions), which are hard to answer from the first and last frames, and \rtqacomp{} (2,544 questions), with two consecutive
events in each composite video.

We evaluate four proprietary and 16 open-source models, with thinking where a model supports it. Every model is more accurate on forward than on reversed videos (\cref{tab:results}). The proprietary models reach 90.6--94.4\% paired
accuracy on \rtqafull, whereas the strongest open-source model, Qwen3.8-27B-Thinking, reaches 66.8\% and most open-source models stay below 35\%. With shuffled frames, the paired accuracy of three tested open-source models drops to 3.4--7.4\%, and
without frames to at most 1.1\% (\cref{tab:controls}). On \rtqacore, the paired accuracy of the proprietary models falls to 55.6--64.5\%, and on \rtqacomp{} they
answer all questions of only 61.0--77.7\% of the groups, the open-source models of at most 26.1\%.

Finally, we construct \rtqatrain{} (\rtqatrainQ{} questions on \rtqatrainV{} videos) from the training split of RTime with the same pipeline, without human verification, and compare supervised fine-tuning (SFT) and the reinforcement learning algorithm DAPO~\cite{yu2025dapo} under the same data, instruction and optimization settings. Training improves all three models, with larger gains from SFT (20.7--27.4 points of paired accuracy) than from DAPO (5.9--22.9 points); for Qwen3.5-9B-Thinking, both methods also yield gains on Vinoground~\cite{zhang2024vinoground}, as does DAPO on TVBench~\cite{cores2024tvbench}.

In summary, our contributions are as follows:
\begin{itemize}
  \item \rtqa, a benchmark of time-reversed question pairs on atomic events, with evaluation sets for fast
  evaluation, intermediate states and two consecutive events;
  \item a construction pipeline that derives questions from event cards, reviewed by a model and verified by humans;
  \item an evaluation of 20 MLLMs with restricted-input controls, revealing lower accuracy on reversed
  videos, a gap between proprietary and open-source models, and a sharp drop in accuracy on questions
  that are hard to answer from the first and last frames;
\item \rtqatrain{} and a comparison of SFT and DAPO under matched settings, in which both improve paired accuracy and also help Qwen3.5-9B-Thinking on other temporal benchmarks.
\end{itemize}

\section{Related Work}
\label{sec:related}

\textbf{Video benchmarks.} Comprehensive benchmarks such as NExT-QA~\cite{xiao2021nextqa},
EgoSchema~\cite{mangalam2023egoschema}, MVBench~\cite{li2024mvbench}, Video-MME~\cite{fu2024videomme} and Perception
Test~\cite{patraucean2023perceptiontest} evaluate MLLMs on causal, temporal and descriptive questions about videos of
increasing length, but analyses show that parts of such benchmarks remain answerable from a single
frame~\cite{buch2022revisiting,lei2022singleframe} or from the text alone~\cite{cores2024tvbench}. \rtqa{} instead focuses on short atomic events and pairs every question with a counterpart whose answer
changes with the frame order.

\textbf{Temporal understanding.} Fine-grained action datasets such as Something-Something~\cite{goyal2017something}
contain classes that differ only in direction, \eg, pushing an object from left to right or from right to left. For
MLLMs, TempCompass~\cite{liu2024tempcompass}, VITATECS~\cite{li2024vitatecs}, TemporalBench~\cite{cai2024temporalbench}
and TOMATO~\cite{shangguan2024tomato} probe order, direction, speed and fine-grained dynamics,
Vinoground~\cite{zhang2024vinoground} contrasts captions that differ only in event order, and Test of
Time~\cite{bagad2023testoftime} examines before and after relations. Classifying whether a video plays forward or
backward is a classic probe of temporal perception~\cite{pickup2014arrow,wei2018arrow}, and recent work shows
that MLLMs often fail at it~\cite{xue2025arrowoftime}. \rtqa{} takes its videos and their reversals from RTime~\cite{rtime}, a video--text retrieval benchmark whose reversed
videos depict different but plausible events, and adds event cards, five question types with identical text and options on
both directions, paired metrics and evaluation sets: unlike the retrieval task of RTime, \rtqa{} asks multiple-choice
questions and credits a pair only if both answers are correct.

\textbf{Reinforcement learning in video language models.} GRPO~\cite{shao2024deepseekmath} and DeepSeek-R1~\cite{guo2025deepseekr1} showed that rule-based rewards elicit
reasoning in language models, and DAPO~\cite{yu2025dapo} and Dr.~GRPO~\cite{liu2025drgrpo} refine its optimization. Video-R1~\cite{feng2025videor1}, VideoChat-R1~\cite{li2025videochatr1} and Time-R1~\cite{wang2025timer1}
apply such rewards to video reasoning, spatio-temporal perception and temporal grounding, and
ArrowRL~\cite{xue2025arrowoftime} rewards responses that differ between a video and its reversal. Comparing
post-training methods, Chu~\etal~\cite{chu2025sft} find that supervised fine-tuning tends to memorize its training data,
whereas reinforcement learning generalizes to unseen variants; we compare DAPO and SFT on identical data.

\section{\rtqa}
\label{sec:rtimeqa}

\subsection{Overview}
\label{sec:overview}

\rtqa{} evaluates whether a multimodal model understands the temporal course of atomic events: the order
of their intermediate states, the step that leads from one state to the next, and the direction in which
the event unfolds. An atomic event is one coherent event with a single start and end, which may consist of several
ordered steps, such as packing toys into a suitcase.

\subsection{Task Formulation}
\label{sec:formulation}

Let $V$ be a video of an atomic event and $\Vr$ the video formed by its frames in reverse order. A question
$q=(V,x,\mathcal{O},a)$ consists of a video, a question text $x$, a set $\mathcal{O}$ of four options and the
correct option $a\in\mathcal{O}$. Each question has a paired question $\qr=(\Vr,x,\mathcal{O},\bar a)$ on
the reversed video, with the same text and options and a different correct option $\bar a\neq a$. A model $f$
receives the frames together with the question text and options, and selects one option $f(q)\in\mathcal{O}$.

\subsection{Benchmark Construction}
\label{sec:construction}

\textbf{Video collection.} We build on the test split of RTime~\cite{rtime}, whose videos were selected for plausible
reversals and are captioned in both directions. Its 994
available forward videos and their reversals form 1,988 candidate videos. We downscale every video to a longer side of at most 640 pixels and encode it with VP9 without audio; the reversed
video, which annotators watch, reverses the frame order before encoding with identical parameters.

\textbf{Event-card annotation.} Rather than writing questions directly, a proposer model (GPT-5.6-Sol)
annotates each forward video with an \emph{event card}, from which all questions are later derived. Given
32 uniformly sampled, timestamped frames and the two RTime captions, the proposer either rejects the video or records
(i) a one-sentence description of the event in $V$ and in $\Vr$; (ii) four key states $s_1,\dots,s_4$ in
chronological order, each with the frames that show it; (iii) for each pair of consecutive states, the
forward step $f_k$ that leads from $s_k$ to $s_{k+1}$ in $V$ and the reverse step $r_k$ that leads from
$s_{k+1}$ back to $s_k$ in $\Vr$; and (iv) distractors, \ie, two states that never occur and two
\emph{path decoys} that share the start and the end of the event but describe a different middle. The
interior states $s_2$ and $s_3$ must lie in the middle 75\% of the video, so that neither the first nor the last frame shows them. A rule-based validator checks these constraints, the ordering of the frame
indices, the distinctness of all entries and the length of the decoys, and returns violations to the
proposer for up to three rounds. Videos without four distinct stages in a fixed order, such as periodic or
back-and-forth motions, are rejected, as are videos with shot cuts and text animations.

\textbf{Question generation.} Templates turn each card into seven question pairs (\textit{supplementary material}): one key-state-order, two
next-state, two next-action, one previous-action and one event-description question, so that every answer can be
checked against its card. Options are shuffled with a fixed seed; the correct answers are balanced across the four
letters (24.7--25.3\%), and correct and incorrect options have similar lengths (10.2--11.7 words).

\textbf{Model review.} A reviewer model (Claude Opus 5) inspects every card and question against the same
frames. It checks that each key state is visible in exactly its listed frames among the 32 sampled ones, that the forward and the
reverse step of each transition are correct, and that every distractor is false for both videos, and it flags questions it deems answerable from a single frame or the text alone. Each question is accepted, revised or rejected;
rejections mostly concern ambiguity or unverifiable claims.

\textbf{Human verification.} Three annotators with higher education and good English proficiency each verified every video retained
by the model review in an interface that plays $V$ and $\Vr$
side by side and shows the event card, thumbnails of the frames assigned to each key state, the questions
and the verdicts of the reviewer. For each video, the annotators (i) evaluated whether the reversed video
matches a meaningful real-world scenario, excluding unrealistic scenarios such as anti-gravity phenomena
or a large number of cars driving backwards on the street; (ii) checked the event card against the frames
of both videos; and (iii) checked every question, including the reviewer's revisions. Majority votes of the
annotators decide each video and question and override the reviewer. Finally, 1,278 of the 1,988 candidate videos remain, with
8,420 questions, \ie, 64.3\% of the candidate videos and 75.5\% of the generated questions; both review stages keep or remove
whole question pairs.

\subsection{Question Types}
\label{sec:types}

\rtqa{} contains five question types (examples in \cref{fig:teaser} and the \textit{supplementary material}); the four options of a question
contain its answers on $V$ and on $\Vr$ and two distractors that are false on both of them.

\textbf{Key-state order} asks for the chronological order of the four key states, listed under shuffled
numbers; its distractors keep the first and last states but swap the interior ones.

\textbf{Next state} asks which state immediately follows an interior state $s_k$; the answer is $s_{k+1}$
on $V$ and $s_{k-1}$ on $\Vr$.

\textbf{Next action} and \textbf{previous action} ask for the step that immediately follows or precedes an
interior state. The answers are forward steps on $V$ and reverse steps on $\Vr$; distractors are
neighboring steps in the wrong direction.

\textbf{Event description} asks for the sentence that describes the whole video, with the two path decoys
as distractors.

\subsection{Statistics and Evaluation Sets}
\label{sec:stats}

\begin{figure}[t]
  \centering
  \includegraphics[width=\linewidth]{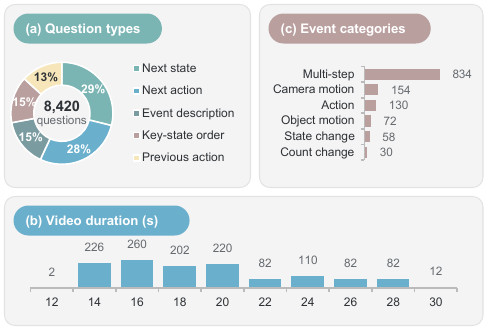}
  \caption{Question types, video durations and event categories.}
  \label{fig:stats}
\end{figure}

\rtqa{} provides four evaluation sets; the \textit{supplementary material} gives their statistics and construction
details. \rtqafull{} contains all 8,420 questions on 1,278 videos, 6.6 questions per video, over the five types of
\cref{fig:stats}a; the videos last 20.2\,s on average (\cref{fig:stats}b), and multi-step events account for 65.3\%
of them (\cref{fig:stats}c). \rtqatiny{} samples 2,036 questions on 304 videos from it, stratified by event category, for fast
evaluation. \rtqacore{} keeps the 428 questions on 204 videos whose pairs neither GPT-5.6-Sol nor Gemini 3.8 Flash
answers from the first and last frames or from the event sentence of the card and that human annotators confirmed to be hard to answer from the first and
last frames alone; the results of these two models on it are therefore selection-conditioned. \rtqacomp{} concatenates two videos of \rtqafull{} in all four direction combinations
(1,272 composite videos, 2,544 questions) and asks for the event of each segment; a group of four composite
videos counts only if all eight answers are correct.

\subsection{Evaluation Protocol}
\label{sec:protocol}

\textbf{Inference.} All models receive the same 32 uniformly sampled frames per video, in reverse order for $\Vr$,
so that the two inputs of a pair differ only in frame order, and $V$ and $\Vr$ are queried separately. Open-source
models run locally and receive one question with its four options per request; proprietary models are queried
through their APIs with all questions on a video in one request, so comparisons across the two groups also reflect
the protocol. Models answer with option letters, answers are extracted by rule, and unanswered questions count as
wrong; the \textit{supplementary material} gives the details and the prompts. We will release the
code, the annotations and the sampled frame indices, subject to the licenses of the RTime source videos.

\textbf{Metrics.} With $c(q)=\mathds{1}[f(q)=a]$, we report
{\small\allowdisplaybreaks
\begin{align}
  &\mathrm{Acc}_V = \frac{1}{|\mathcal{Q}^{+}|}\sum_{q\in\mathcal{Q}^{+}} c(q), \label{eq:acc}\\
  &\mathrm{Acc}_{\mathrm{pair}} = \frac{1}{|\mathcal{Q}^{+}|}\sum_{q\in\mathcal{Q}^{+}} c(q)\,c(\qr), \label{eq:pair}\\
  &\mathrm{Acc}_{\mathrm{video}} = \frac{1}{|\mathcal{V}|}\sum_{v\in\mathcal{V}}\,\prod_{q\in\mathcal{Q}_v} c(q), \label{eq:video}\\
  &\mathrm{Acc}_{\mathrm{strict}} = \frac{1}{|\mathcal{V}^{+}|}\sum_{v\in\mathcal{V}^{+}}\,\prod_{q\in\mathcal{Q}_v} c(q)\,c(\qr), \label{eq:strict}
\end{align}}
where $\mathcal{Q}^{+}$ denotes the questions on forward videos, $\mathcal{V}$ all videos,
$\mathcal{V}^{+}\subset\mathcal{V}$ the forward videos, and $\mathcal{Q}_v$ the questions on video $v$;
$\mathrm{Acc}_{\Vr}$ is defined as $\mathrm{Acc}_V$ over the paired questions $\qr$. The paired accuracy
and the strict accuracy are the counterparts of the question accuracy and the video accuracy that also
require the reversed video to be answered correctly; uniform guessing yields a paired accuracy of 6.25\%. Paired accuracy on \rtqafull{} is the main metric; on \rtqacore, the video-level metrics consider only the questions of each video that belong to \rtqacore.

\section{Learning Atomic Temporal Events}
\label{sec:rl}

We next examine whether the temporal understanding measured by \rtqa{} can be learned from data of the
same kind. We build \rtqatrain{} (\cref{sec:rtqatrain}) and compare two ways of learning from it: reinforcement learning
(RL) with DAPO~\cite{yu2025dapo}, following RL approaches to temporal video understanding~\cite{wang2025timer1}
(\cref{sec:grpo}), and supervised fine-tuning (SFT) on the correct options (\cref{sec:sft}).

\subsection{\rtqatrain}
\label{sec:rtqatrain}

\rtqatrain{} is built from the training split of RTime~\cite{rtime}, which shares no videos with the test split. From
its 6,745 videos with a meaningful reversal, we sample 1,000 uniformly at random, add
their reversals (2,000 videos) and apply the
construction pipeline of \cref{sec:construction} except for human verification. \rtqatrain{} comprises \rtqatrainQ{} questions on the \rtqatrainV{} videos that pass the pipeline. We hold
out 24 video pairs (326 questions on 48 videos) for validation and train on the remaining 8,640 questions on 1,304 videos. The two questions of a pair enter
training as separate samples.

\begin{table*}[t]
  \centering
  \small
  \setlength{\tabcolsep}{3pt}
  \caption{Results on \rtqafull, \rtqatiny{} and \rtqacore{} (8,420, 2,036 and 428 questions); metrics as in \cref{eq:acc,eq:pair,eq:video,eq:strict}, with Pair as the
  main metric. Random: expected value of uniform guessing. Shading: top three per group and column, darker for better.}
  \label{tab:results}
  \begin{tabular*}{\linewidth}{@{\extracolsep{\fill}}lrrrrrrrrrrrrrrr}
    \toprule
    & \multicolumn{5}{c}{\rtqafull} & \multicolumn{5}{c}{\rtqatiny} & \multicolumn{5}{c}{\rtqacore} \\
    \cmidrule(lr){2-6}\cmidrule(lr){7-11}\cmidrule(lr){12-16}
    Model & $V$ & $\Vr$ & Pair & Video & Strict & $V$ & $\Vr$ & Pair & Video & Strict & $V$ & $\Vr$ & Pair & Video & Strict \\
    \midrule
    Random & 25.0 & 25.0 & 6.3 & 0.5 & 0.1 & 25.0 & 25.0 & 6.3 & 0.2 & 0.0 & 25.0 & 25.0 & 6.3 & 16.7 & 4.0 \\
    \midrule
    \multicolumn{16}{l}{\textit{Proprietary Models (API)}} \\
    Gemini 3.1 Pro & \ranktwo{98.6} & \ranktwo{93.7} & \ranktwo{92.4} & \ranktwo{89.4} & \ranktwo{80.8} & \ranktwo{99.1} & \ranktwo{95.7} & \ranktwo{94.8} & \ranktwo{92.8} & \ranktwo{86.2} & \rankone{95.3} & \rankthree{65.4} & \rankthree{61.7} & \rankone{77.9} & \rankone{56.9} \\
    GPT-5.6-Sol & \rankthree{98.5} & 92.0 & 90.6 & 84.0 & 71.8 & \rankthree{98.7} & \rankthree{93.8} & \rankthree{92.6} & \rankthree{87.5} & \rankthree{76.3} & \rankthree{93.5} & 60.7 & 55.6 & \ranktwo{75.0} & \rankthree{53.9} \\
    Claude Opus 5 & \rankthree{98.5} & \rankthree{93.3} & \rankthree{92.1} & \rankthree{86.2} & \rankthree{75.3} & \rankone{99.2} & 92.1 & 91.6 & 87.2 & \rankthree{76.3} & \rankthree{93.5} & \rankone{67.8} & \ranktwo{64.0} & \ranktwo{75.0} & \rankone{56.9} \\
    Gemini 3.8 Flash & \rankone{98.9} & \rankone{95.3} & \rankone{94.4} & \rankone{89.7} & \rankone{82.5} & \ranktwo{99.1} & \rankone{96.4} & \rankone{95.7} & \rankone{93.1} & \rankone{87.5} & \ranktwo{94.9} & \ranktwo{66.8} & \rankone{64.5} & \rankthree{74.5} & \ranktwo{54.9} \\
    \midrule
    \multicolumn{16}{l}{\textit{Open-source Models}} \\
    LLaVA-OneVision-7B & 46.1 & 34.0 & 10.9 & 1.0 & 0.0 & 45.1 & 36.9 & 11.2 & 1.0 & 0.0 & 45.8 & 24.8 & 8.4 & 23.0 & 2.9 \\
    LLaVA-Video-7B & 53.3 & 41.5 & 19.0 & 2.7 & 0.5 & 52.4 & 44.7 & 19.0 & 2.3 & 0.7 & 45.3 & 32.7 & 9.3 & 27.0 & 4.9 \\
    VideoLLaMA3-7B & 36.0 & 29.6 & 1.5 & 0.5 & 0.0 & 35.7 & 29.5 & 0.9 & 0.3 & 0.0 & 33.2 & 22.9 & 1.4 & 17.6 & 2.0 \\
    Cosmos-Reason1-7B & 50.9 & 37.5 & 17.5 & 1.8 & 0.0 & 51.1 & 40.6 & 18.3 & 2.3 & 0.0 & 43.9 & 25.7 & 8.4 & 24.0 & 5.9 \\
    GLM-4.1V-9B-Thinking & 60.0 & 48.5 & 26.2 & 4.9 & 0.5 & 58.5 & 50.2 & 26.3 & 3.6 & 0.7 & 52.3 & 34.1 & 12.6 & 30.4 & 10.8 \\
    MiniCPM-V-4.5 & 57.6 & 44.2 & 23.0 & 5.2 & 0.8 & 57.2 & 45.6 & 23.1 & 5.6 & 0.7 & 48.6 & 31.3 & 11.7 & 29.9 & 8.8 \\
    InternVL3.5-8B-Thinking & 62.7 & 50.5 & 30.6 & 5.9 & 0.8 & 61.5 & 52.0 & 30.5 & 3.6 & 1.3 & 57.0 & 36.4 & 16.4 & 35.3 & 13.7 \\
    InternVL3.5-14B-Thinking & 65.6 & 54.2 & 34.1 & 9.6 & 2.0 & 63.8 & 57.6 & 35.5 & 8.2 & 0.0 & 56.5 & 38.8 & 16.4 & 35.3 & 16.7 \\
    Keye-VL-1.5-8B & 51.4 & 41.2 & 16.9 & 2.1 & 0.0 & 48.9 & 43.5 & 17.1 & 1.3 & 0.0 & 45.3 & 28.5 & 3.7 & 25.0 & 0.0 \\
    Qwen3-VL-8B-Instruct & 65.4 & 52.5 & 33.0 & 8.9 & 1.6 & 62.8 & 56.0 & 33.0 & 6.6 & 2.0 & 62.6 & 36.9 & 16.4 & 39.7 & 16.7 \\
    Qwen3-VL-8B-Thinking & 68.3 & 53.0 & 34.9 & 8.8 & 0.9 & 66.3 & 56.0 & 36.1 & 8.2 & 0.7 & 60.7 & 37.4 & 19.2 & 38.7 & 16.7 \\
    Molmo2-8B & 61.7 & 50.6 & 28.8 & 4.3 & 0.5 & 61.0 & 53.7 & 30.6 & 3.3 & 0.7 & 59.3 & 33.2 & 15.9 & 34.8 & 13.7 \\
    Qwen3.5-9B-Thinking & 72.9 & 58.2 & 43.1 & 12.7 & 2.8 & 73.4 & 63.1 & 47.3 & 11.8 & 2.0 & 63.1 & 36.4 & 24.3 & 37.3 & 18.6 \\
    Qwen3.5-27B-Thinking & \ranktwo{83.4} & \ranktwo{70.8} & \ranktwo{59.8} & \ranktwo{28.6} & \ranktwo{11.0} & \ranktwo{83.2} & \ranktwo{73.8} & \ranktwo{62.0} & \rankthree{26.0} & \rankthree{8.6} & \ranktwo{76.6} & \ranktwo{47.2} & \rankthree{34.1} & \ranktwo{49.5} & \ranktwo{26.5} \\
    Qwen3.6-27B-Thinking & \rankthree{81.2} & \rankthree{67.8} & \rankthree{56.3} & \rankthree{27.2} & \rankthree{10.3} & \rankthree{81.8} & \rankthree{69.6} & \rankthree{57.7} & \ranktwo{26.6} & \ranktwo{9.2} & \rankthree{70.6} & \rankthree{44.9} & \ranktwo{35.5} & \rankthree{45.6} & \rankthree{24.5} \\
    Qwen3.8-27B-Thinking & \rankone{86.5} & \rankone{75.6} & \rankone{66.8} & \rankone{36.7} & \rankone{19.4} & \rankone{87.0} & \rankone{77.4} & \rankone{68.9} & \rankone{36.2} & \rankone{15.8} & \rankone{78.0} & \rankone{56.5} & \rankone{44.4} & \rankone{57.4} & \rankone{38.2} \\
    \bottomrule
  \end{tabular*}
\end{table*}

\subsection{Reinforcement Learning with DAPO}
\label{sec:grpo}

\textbf{Objective.} Given a question $q$ and the frames of its video, DAPO samples a group
of $G$ responses $\{o_1,\ldots,o_G\}$ from the current policy $\pi_{\theta_{\mathrm{old}}}$ and scores each response
with a reward $r_i=r(o_i)$. Instead of a learned value function, the advantage of a response is its reward minus
the group mean, without division by the standard deviation~\cite{liu2025drgrpo}, shared by all its tokens. The policy $\pi_\theta$ maximizes a token-level clipped surrogate objective with
decoupled clipping ranges and no KL penalty~\cite{yu2025dapo}:
{\small
\begin{align}
  &A_i = r_i-\operatorname{mean}\big(\{r_j\}_{j=1}^{G}\big), \label{eq:adv}\\
  &\mathcal{J}(\theta) = \mathbb{E}\Big[\frac{1}{\sum_{i=1}^{G}|o_i|}\sum_{i=1}^{G}\sum_{t=1}^{|o_i|}
    \min\big(\rho_{i,t} A_i,\nonumber\\
  &\phantom{\mathcal{J}(\theta) =}\;\operatorname{clip}(\rho_{i,t},1{-}\epsilon_{\mathrm{low}},1{+}\epsilon_{\mathrm{high}})\,A_i\big)\Big], \label{eq:grpo}
\end{align}}
where $\rho_{i,t}=\pi_\theta(o_{i,t}\mid q,o_{i,<t})/\pi_{\theta_{\mathrm{old}}}(o_{i,t}\mid q,o_{i,<t})$ is the
probability ratio and $\epsilon_{\mathrm{low}}<\epsilon_{\mathrm{high}}$ the clipping ranges.

\textbf{Reward.} The reward combines an accuracy reward for the chosen option with a format reward:
{\small
\begin{align}
  &r_{\mathrm{acc}}(o) = \mathds{1}\big[\hat{a}(o)=a\big], \label{eq:racc}\\
  &r_{\mathrm{form}}(o) = \mathds{1}\big[o\in\mathcal{T}\big], \label{eq:rform}\\
  &r(o) = 0.9\,r_{\mathrm{acc}}(o) + 0.1\,r_{\mathrm{form}}(o), \label{eq:reward}
\end{align}}
where $\hat{a}(o)$ is the option stated in the answer of response $o$, $a$ the correct option on the queried video
($\bar{a}$ for a question $\qr$ on $\Vr$), and $\mathcal{T}$ the set of responses that follow the template
``\texttt{<think>}$\,\cdots\,$\texttt{</think><answer>}$\,\cdots\,$\texttt{</answer>}'' with a single option letter
as the answer. The format reward enforces a parseable response with a separate reasoning part but does not assess
its content; the accuracy reward credits only the correct option of the queried direction.

\textbf{Training.} We train for one epoch with verl~\cite{sheng2024hybridflow} and evaluate with the training instruction;
the \textit{supplementary material} lists the models, prompts and hyperparameters.

\subsection{Supervised Fine-Tuning}
\label{sec:sft}

As a baseline that learns from the same questions without sampling, SFT trains the model to output the correct
option: given the frames and the training instruction, it minimizes the negative log-likelihood
{\small
\begin{equation}
  \mathcal{L}_{\mathrm{SFT}}(\theta) = -\mathbb{E}_{q}\Big[\sum_{t=1}^{|y|}\log \pi_\theta\big(y_t \mid v, x, \mathcal{O}, y_{<t}\big)\Big],
  \label{eq:sft}
\end{equation}}
where $v\in\{V,\Vr\}$ is the queried video and the target $y$ is an empty reasoning block followed by its correct
option, since \rtqatrain{} contains no reasoning; only the tokens of $y$ enter the loss, so the model learns to answer without reasoning. SFT
fine-tunes all parameters for one epoch, and the trained models are evaluated with the training instruction
(\textit{supplementary material}).

\section{Experiments}
\label{sec:experiments}

\subsection{Setup}
\label{sec:exp_setup}

\textbf{Models.} We evaluate four proprietary models, Gemini 3.1 Pro-20260219~\cite{gemini31pro}, GPT-5.6-Sol-20260709~\cite{gpt56}, Claude Opus 5-20260724~\cite{claudeopus5} and Gemini 3.8 Flash-20260902~\cite{gemini38flash} (suffixes give the release date of the
evaluated version), and sixteen open-source models,
LLaVA-OneVision-7B~\cite{li2024llavaonevision}, LLaVA-Video-7B~\cite{zhang2024llavavideo}, VideoLLaMA3-7B~\cite{zhang2025videollama3}, Cosmos-Reason1-7B~\cite{azzolini2025cosmosreason1}, GLM-4.1V-9B-Thinking~\cite{hong2025glm41v}, MiniCPM-V-4.5~\cite{yu2025minicpmv45}, InternVL3.5-8B and InternVL3.5-14B~\cite{wang2025internvl35}, Keye-VL-1.5-8B~\cite{yang2025keyevl15}, Qwen3-VL-8B-Instruct and Qwen3-VL-8B-Thinking~\cite{bai2025qwen3vl}, Molmo2-8B~\cite{clark2026molmo2}, Qwen3.5-9B and Qwen3.5-27B~\cite{qwen35}, Qwen3.6-27B~\cite{qwen36}, and Qwen3.8-27B~\cite{qwen38}. InternVL3.5-8B, InternVL3.5-14B, Qwen3.5-9B, Qwen3.5-27B, Qwen3.6-27B and Qwen3.8-27B support a thinking and a non-thinking mode and are reported with thinking (InternVL3.5 through the thinking system prompt of its model card), marked by the suffix -Thinking in the tables, and Cosmos-Reason1-7B, GLM-4.1V-9B-Thinking and Qwen3-VL-8B-Thinking always reason before answering, which yields 20 configurations.
 Two of the proprietary models, GPT-5.6-Sol and Claude Opus 5, also served as the proposer and the
reviewer during construction (\cref{sec:construction}), so their results are not independent of it; the two Gemini
models took no part in the construction, although Gemini 3.8 Flash helped select \rtqacore{} (\cref{sec:stats}).

\textbf{Evaluation.} The \textit{supplementary material} also gives restricted-input baselines of proprietary
models and 95\% bootstrap confidence intervals.

\subsection{Main Results}
\label{sec:exp_main}

\begin{table}[t]
  \centering
  \small
  \setlength{\tabcolsep}{2.2pt}
  \caption{Paired accuracy per question type on \rtqafull: key-state order (Key. Ord.), next state (Next St.), next action (Next Act.), previous action (Prev. Act.), event description (Evt. Desc.). Shading: top three per group and column, darker for better.}
  \label{tab:results_types}
  \begin{tabular*}{\linewidth}{@{\extracolsep{\fill}}lrrrrrr}
    \toprule
    Model & \shortstack{Key.\\Ord.} & \shortstack{Next\\St.} & \shortstack{Next\\Act.} & \shortstack{Prev.\\Act.} & \shortstack{Evt.\\Desc.} & All \\
    \midrule
    Random & 6.3 & 6.3 & 6.3 & 6.3 & 6.3 & 6.3 \\
    \midrule
    \multicolumn{7}{l}{\textit{Proprietary Models (API)}} \\
    Gemini 3.1 Pro & \ranktwo{93.1} & \rankthree{93.3} & \rankthree{92.0} & \ranktwo{89.2} & \rankthree{93.2} & \ranktwo{92.4} \\
    GPT-5.6-Sol & \rankthree{92.3} & 92.0 & 89.1 & 85.9 & \ranktwo{93.4} & 90.6 \\
    Claude Opus 5 & 90.4 & \ranktwo{94.5} & \rankone{92.9} & \rankthree{86.7} & 92.8 & \rankthree{92.1} \\
    Gemini 3.8 Flash & \rankone{96.3} & \rankone{96.5} & \ranktwo{92.6} & \rankone{90.6} & \rankone{95.1} & \rankone{94.4} \\
    \midrule
    \multicolumn{7}{l}{\textit{Open-source Models}} \\
    LLaVA-OneVision-7B & 17.8 & 8.1 & 9.1 & 9.4 & 14.5 & 10.9 \\
    LLaVA-Video-7B & 24.1 & 18.9 & 20.1 & 11.4 & 19.2 & 19.0 \\
    VideoLLaMA3-7B & 2.3 & 1.2 & 1.8 & 1.8 & 0.8 & 1.5 \\
    Cosmos-Reason1-7B & 19.9 & 16.0 & 14.9 & 16.0 & 24.1 & 17.5 \\
    GLM-4.1V-9B-Thinking & 40.4 & 19.9 & 27.5 & 15.5 & 31.3 & 26.2 \\
    MiniCPM-V-4.5 & 30.1 & 19.8 & 24.5 & 17.8 & 24.2 & 23.0 \\
    InternVL3.5-8B-Thinking & 39.7 & 23.5 & 33.7 & 17.3 & 40.9 & 30.6 \\
    InternVL3.5-14B-Thinking & 49.4 & 28.2 & 30.0 & 24.7 & 46.4 & 34.1 \\
    Keye-VL-1.5-8B & 24.8 & 14.0 & 15.2 & 11.0 & 23.3 & 16.9 \\
    Qwen3-VL-8B-Instruct & 46.1 & 27.4 & 30.8 & 24.5 & 42.6 & 33.0 \\
    Qwen3-VL-8B-Thinking & 51.9 & 28.1 & 30.9 & 26.3 & 45.9 & 34.9 \\
    Molmo2-8B & 45.7 & 23.5 & 29.4 & 15.0 & 33.2 & 28.8 \\
    Qwen3.5-9B-Thinking & 55.6 & 38.2 & 43.2 & 32.4 & 49.2 & 43.1 \\
    Qwen3.5-27B-Thinking & \rankone{79.7} & \ranktwo{52.4} & \ranktwo{60.2} & \ranktwo{50.5} & \ranktwo{62.1} & \ranktwo{59.8} \\
    Qwen3.6-27B-Thinking & \rankthree{73.6} & \rankthree{50.2} & \rankthree{58.2} & \rankthree{42.7} & \rankthree{59.6} & \rankthree{56.3} \\
    Qwen3.8-27B-Thinking & \ranktwo{79.4} & \rankone{60.7} & \rankone{67.2} & \rankone{54.6} & \rankone{75.9} & \rankone{66.8} \\
    \bottomrule
  \end{tabular*}
\end{table}

\textbf{Reversed videos are harder for every model.} The proprietary models answer 98.5--98.9\% of the questions on
$V$ but 92.0--95.3\% on $\Vr$, the open-source models 36.0--86.5\% and 29.6--75.6\% (\cref{tab:results}).

\textbf{Paired accuracy separates models with similar forward accuracy.} The proprietary models differ by at
most 0.4 points in $\mathrm{Acc}_V$ but by 3.8 in paired accuracy, from 90.6\%
(GPT-5.6-Sol) to 94.4\% (Gemini 3.8 Flash). The best open-source model, Qwen3.8-27B-Thinking, reaches
66.8\%, 23.8 points below the weakest proprietary model.

\textbf{Answering all questions on a video remains difficult.} Gemini 3.8 Flash answers all questions on 89.7\% of the videos
and on both videos of 82.5\% of the video pairs, whereas no open-source model exceeds 36.7\%
video accuracy or 19.4\% strict accuracy.


\textbf{Most open-source models rarely answer both directions.} Apart from the Qwen3.5, Qwen3.6 and Qwen3.8 models,
the open-source models reach only 1.5--34.9\% paired accuracy, against 6.3\% for random guessing, although they
answer 36.0--68.3\% of the questions on forward videos. This includes models built for video understanding,
LLaVA-Video-7B, VideoLLaMA3-7B and Molmo2-8B, and the reasoning models Cosmos-Reason1-7B, GLM-4.1V-9B-Thinking and Qwen3-VL-8B-Thinking. None of them answers all questions on both videos
of more than 2.0\% of the video pairs, and VideoLLaMA3-7B gives the same answer on both videos for
92.5\% of the questions, although the correct answers always differ.

\textbf{\rtqacore{} exposes a large gap.} On \rtqacore, paired accuracy decreases by 28.1--35.0 points for
the proprietary models and by up to 25.7 points for the open-source models. The decrease concentrates on
reversed videos: the proprietary models still answer 93.5--95.3\% on forward videos but only 60.7--67.8\% on reversed
videos.

\textbf{Previous-action questions are the hardest.} Previous action yields the lowest paired accuracy of the
five question types for 17 of the 20 configurations (\cref{tab:results_types}). The proprietary models span
85.9--96.5\% across types, while the open-source models reach 2.3--79.7\% on
key-state order, their easiest type, and 1.8--54.6\% on previous action.

\begin{table}[t]
  \centering
  \small
  \setlength{\tabcolsep}{3pt}
  \caption{\rtqacomp. Video: both questions of a composite
  video correct; Group: all eight questions of a group correct. Shading: top three per group and column, darker for better.}
  \label{tab:results_composite}
  \begin{tabular*}{\linewidth}{@{\extracolsep{\fill}}lrrr}
    \toprule
    Model & Question & Video & Group \\
    \midrule
    Random & 25.0 & 6.3 & 0.0 \\
    \midrule
    \multicolumn{4}{l}{\textit{Proprietary Models (API)}} \\
    Gemini 3.1 Pro & 90.6 & 82.5 & 61.0 \\
    GPT-5.6-Sol & \ranktwo{93.8} & \ranktwo{88.1} & \ranktwo{74.2} \\
    Claude Opus 5 & \rankthree{92.3} & \rankthree{85.1} & \rankthree{67.6} \\
    Gemini 3.8 Flash & \rankone{94.8} & \rankone{90.2} & \rankone{77.7} \\
    \midrule
    \multicolumn{4}{l}{\textit{Open-source Models}} \\
    LLaVA-OneVision-7B & 36.1 & 12.7 & 0.3 \\
    LLaVA-Video-7B & 37.8 & 14.4 & 2.5 \\
    VideoLLaMA3-7B & 29.6 & 9.3 & 0.3 \\
    Cosmos-Reason1-7B & 41.9 & 17.6 & 0.0 \\
    GLM-4.1V-9B-Thinking & 51.2 & 26.2 & 4.4 \\
    MiniCPM-V-4.5 & 43.2 & 18.1 & 3.5 \\
    InternVL3.5-8B-Thinking & 55.0 & 30.0 & 7.2 \\
    InternVL3.5-14B-Thinking & 56.8 & 31.7 & 7.9 \\
    Keye-VL-1.5-8B & 40.2 & 16.2 & 1.6 \\
    Qwen3-VL-8B-Instruct & 59.6 & 35.5 & 10.1 \\
    Qwen3-VL-8B-Thinking & 59.1 & 34.2 & 6.3 \\
    Molmo2-8B & 51.1 & 25.9 & 3.8 \\
    Qwen3.5-9B-Thinking & 60.8 & 37.0 & 9.1 \\
    Qwen3.5-27B-Thinking & \ranktwo{67.0} & \ranktwo{44.0} & \ranktwo{17.0} \\
    Qwen3.6-27B-Thinking & \rankthree{66.3} & \rankthree{43.9} & \rankthree{15.4} \\
    Qwen3.8-27B-Thinking & \rankone{77.7} & \rankone{59.4} & \rankone{26.1} \\
    \bottomrule
  \end{tabular*}
\end{table}

\textbf{Group accuracy on \rtqacomp{} is low.} The proprietary models answer 90.6--94.8\% of the questions
but only 61.0--77.7\% of the groups (\cref{tab:results_composite}); the open-source models reach 29.6--77.7\% and 0.0--26.1\%,
respectively.

\subsection{Analysis}
\label{sec:exp_analysis}

\textbf{The tested open-source models depend on frame order.} With the 32 frames in a random order, the paired accuracy of these models falls from 34.1--66.8\% to 3.4--7.4\% (\cref{tab:controls}), and on \rtqacore{} from 16.4--44.4\% to 0.9--7.9\%. Without frames, it is 0.4--1.1\%, although they answer 42.2--51.5\% of the questions on forward videos and only 28.7--30.4\% on reversed ones: language priors favor the forward answer. With only the first and last frames, it is 4.8--18.2\%, well below their accuracy with all 32 frames.

\subsection{Learning from \rtqatrain}
\label{sec:exp_rl}

\textbf{Training on \rtqatrain{} improves paired accuracy.} Compared with the untrained models, SFT raises the paired accuracy of the three models on \rtqafull{} by 20.7--27.4 points, and DAPO raises it by 5.9--22.9 points (\cref{tab:rl}). On \rtqacore, SFT raises paired accuracy by 15.4--20.1 points, and DAPO raises it by 1.8--19.2 points. SFT exceeds DAPO for every model, by 4.5--16.7 points on \rtqafull, although DAPO keeps the checkpoint with the best validation accuracy and SFT its final one. SFT and DAPO share the training prompt and the baselines use the standard prompt, so the SFT--DAPO comparison is prompt-matched, whereas the gains over the baselines include the prompt change (\textit{supplementary material}). As the test videos share the templates of \rtqatrain{} and each setting is trained once, this comparison does not establish which method generalizes better under broader shifts. The \textit{supplementary material} shows example outputs.

\textbf{Ablations.} With the accuracy reward alone, DAPO brings the paired accuracy of Qwen3-VL-8B-Thinking to 41.5\% on \rtqafull{} and 22.9\% on \rtqacore{}, against 40.8\% and 22.4\% with both rewards (\cref{tab:ablation}).
\begin{table}[t]
  \centering
  \small
  \setlength{\tabcolsep}{3pt}
  \caption{Restricted-input controls. First+last: only the first and the last frame; shuffled: the 32 frames in a fixed random order per question and direction, with timestamps assigned by position. Core: paired accuracy on \rtqacore.}
  \label{tab:controls}
  \begin{tabular*}{\linewidth}{@{\extracolsep{\fill}}llrrrr}
    \toprule
    & & \multicolumn{3}{c}{\rtqafull} & Core \\
    \cmidrule(lr){3-5}\cmidrule(lr){6-6}
    Model & Input & $V$ & $\Vr$ & Pair & Pair \\
    \midrule
    InternVL3.5-14B-Thinking & No frames & 42.2 & 28.7 & 0.6 & 0.5 \\
     & First+last & 50.3 & 38.5 & 13.6 & 3.7 \\
     & Shuffled & 36.2 & 28.6 & 3.4 & 0.9 \\
     & Ordered & 65.6 & 54.2 & 34.1 & 16.4 \\
    \midrule
    Qwen3-VL-8B-Thinking & No frames & 43.2 & 29.6 & 0.4 & 0.9 \\
     & First+last & 45.6 & 29.5 & 4.8 & 2.3 \\
     & Shuffled & 41.8 & 27.8 & 4.4 & 6.1 \\
     & Ordered & 68.3 & 53.0 & 34.9 & 19.2 \\
    \midrule
    Qwen3.8-27B-Thinking & No frames & 51.5 & 30.4 & 1.1 & 0.0 \\
     & First+last & 55.7 & 39.5 & 18.2 & 8.9 \\
     & Shuffled & 41.0 & 29.0 & 7.4 & 7.9 \\
     & Ordered & 86.5 & 75.6 & 66.8 & 44.4 \\
    \bottomrule
  \end{tabular*}
\end{table}
\begin{table}[t]
  \centering
  \small
  \setlength{\tabcolsep}{3pt}
  \caption{Training on \rtqatrain. Baseline: prompt of \cref{sec:protocol}; SFT and
  DAPO share the training prompt and settings. Core: paired accuracy on \rtqacore; Comp.: group accuracy on
  \rtqacomp. Shading: best per model.}
  \label{tab:rl}
  \begin{tabular*}{\linewidth}{@{\extracolsep{\fill}}lrrrrrr}
    \toprule
    & \multicolumn{4}{c}{\rtqafull} & Core & Comp. \\
    \cmidrule(lr){2-5}
    Model & $V$ & $\Vr$ & Pair & Strict & Pair & Group \\
    \midrule
    \multicolumn{7}{l}{\textit{Qwen3-VL-8B-Thinking}} \\
    Baseline & 68.3 & 53.0 & 34.9 & 0.9 & 19.2 & 6.3 \\
    \quad + SFT & \rankone{78.0} & \rankone{71.5} & \rankone{57.5} & \rankone{8.8} & \rankone{37.9} & \rankone{36.5} \\
    \quad + DAPO & 72.2 & 57.7 & 40.8 & 2.7 & 22.4 & 16.4 \\
    \midrule
    \multicolumn{7}{l}{\textit{Qwen3.5-9B-Thinking}} \\
    Baseline & 72.9 & 58.2 & 43.1 & 2.8 & 24.3 & 9.1 \\
    \quad + SFT & \rankone{85.8} & \rankone{80.3} & \rankone{70.5} & \rankone{23.6} & \rankone{44.4} & \rankone{54.7} \\
    \quad + DAPO & 84.9 & 75.8 & 66.0 & 19.6 & 43.5 & 37.7 \\
    \midrule
    \multicolumn{7}{l}{\textit{InternVL3.5-8B-Thinking}} \\
    Baseline & 62.7 & 50.5 & 30.6 & 0.8 & 16.4 & 7.2 \\
    \quad + SFT & \rankone{73.9} & \rankone{67.0} & \rankone{51.3} & \rankone{5.9} & \rankone{31.8} & \rankone{25.8} \\
    \quad + DAPO & 68.4 & 55.5 & 38.2 & 1.9 & 18.2 & 8.8 \\
    \bottomrule
  \end{tabular*}
\end{table}
\begin{table}[t]
  \centering
  \small
  \setlength{\tabcolsep}{3pt}
  \caption{DAPO ablations on Qwen3-VL-8B-Thinking. Accuracy reward only: $r_{\mathrm{acc}}$ without
  $r_{\mathrm{form}}$; forward videos only: the 4,320 questions on forward videos for two epochs. Columns as
  in \cref{tab:rl}.}
  \label{tab:ablation}
  \begin{tabular*}{\linewidth}{@{\extracolsep{\fill}}lrrrrr}
    \toprule
    & \multicolumn{3}{c}{\rtqafull} & Core & Comp. \\
    \cmidrule(lr){2-4}
    Training & $V$ & $\Vr$ & Pair & Pair & Group \\
    \midrule
    Baseline & 68.3 & 53.0 & 34.9 & 19.2 & 6.3 \\
    DAPO & 72.2 & \rankone{57.7} & 40.8 & 22.4 & 16.4 \\
    Accuracy reward only & 72.6 & \rankone{57.7} & 41.5 & 22.9 & 16.7 \\
    Forward videos only & \rankone{73.3} & 57.6 & \rankone{41.7} & \rankone{24.3} & \rankone{17.0} \\
    \bottomrule
  \end{tabular*}
\end{table}
Training only on forward videos yields 41.7\% and 24.3\%, with 57.6\% accuracy on reversed videos against 57.7\% when the reversed videos are part of the training data. The three variants differ by at most 0.9 points in paired accuracy on \rtqafull{} and 1.9 points on \rtqacore. The \textit{supplementary material} plots the validation accuracy and the response length of the three runs over training.

\begin{table}[t]
  \centering
  \small
  \setlength{\tabcolsep}{3pt}
  \caption{Qwen3.5-9B-Thinking on two other temporal benchmarks, Vinoground~\cite{zhang2024vinoground} and TVBench~\cite{cores2024tvbench}, before and after training on \rtqatrain. Shading: best per column.}
  \label{tab:transfer}
  \begin{tabular*}{\linewidth}{@{\extracolsep{\fill}}lrrrr}
    \toprule
    & \multicolumn{2}{c}{Vinoground} & \multicolumn{2}{c}{TVBench} \\
    \cmidrule(lr){2-3}\cmidrule(lr){4-5}
    Model & Acc. & Text & Macro & Micro \\
    \midrule
    \multicolumn{5}{l}{\textit{Qwen3.5-9B-Thinking}} \\
    Baseline & 84.2 & 71.2 & 65.7 & 64.0 \\
    \quad + SFT & \rankone{87.5} & \rankone{77.0} & 65.8 & 64.0 \\
    \quad + DAPO & 85.8 & 75.0 & \rankone{70.3} & \rankone{69.5} \\
    \bottomrule
  \end{tabular*}
\end{table}

\textbf{\rtqatrain{} also helps on other temporal benchmarks.} We evaluate Qwen3.5-9B-Thinking on two other temporal benchmarks. Vinoground~\cite{zhang2024vinoground} pairs captions that use the same words in a different temporal order: each video asks which caption of its pair describes it, and the text score credits a pair if both videos are answered correctly. TVBench~\cite{cores2024tvbench}, designed against single-frame and text-only shortcuts, is evaluated on nine of its ten tasks (2,205 questions; the NTU RGB+D clips of Action Antonym are distributed separately) with the mean task accuracy (macro) and the accuracy over all questions (micro). The Vinoground text score rises by 5.8 points after SFT and by 3.8 points after DAPO, and question accuracy by 3.3 points and by 1.6 points, respectively (\cref{tab:transfer}). On TVBench, the macro accuracy is nearly unchanged after SFT ($+0.1$ points) and rises by 4.6 points after DAPO.

\section{Conclusion}
\label{sec:conclusion}

We presented \rtqa, a benchmark that evaluates atomic temporal event understanding with time-reversed question pairs.
Paired questions neutralize order-invariant language priors: the three tested open-source models answer at
most 1.1\% of the pairs without frames and at most 7.4\% with shuffled frames. All 20
evaluated models answer forward videos more accurately than reversed ones. Under our inference protocols, the
open-source models trail the proprietary models by more than 20 points in paired accuracy; all models score lower on
\rtqacore, whose questions are hard to answer from the first and last frames, and few answer all questions of a \rtqacomp{}
group.
With matched settings, SFT on \rtqatrain{} improves every model by up to 27.4 points and DAPO by up to 22.9 points; for Qwen3.5-9B-Thinking, both methods also improve the Vinoground text score, and DAPO additionally improves TVBench accuracy.



{
    \small
    \bibliographystyle{ieeenat_fullname}
    \bibliography{main}

@String(CVPR= {IEEE Conf. Comput. Vis. Pattern Recog.})

@String(ICCV= {Int. Conf. Comput. Vis.})

@String(CVPR  = {CVPR})

@String(ICCV  = {ICCV})

@inproceedings{rtime,
  title={Reversed in time: A novel temporal-emphasized benchmark for cross-modal video-text retrieval},
  author={Du, Yang and Liu, Yuqi and Jin, Qin},
  booktitle={Proceedings of the 32nd ACM International Conference on Multimedia},
  pages={5260--5269},
  year={2024}
}

@article{shao2024deepseekmath,
  title={Deepseekmath: Pushing the limits of mathematical reasoning in open language models},
  author={Shao, Zhihong and Wang, Peiyi and Zhu, Qihao and Xu, Runxin and Song, Junxiao and Bi, Xiao and Zhang, Haowei and Zhang, Mingchuan and Li, YK and Wu, Yang and others},
  journal={arXiv preprint arXiv:2402.03300},
  year={2024}
}

@article{guo2025deepseekr1,
  title={Deepseek-r1: Incentivizing reasoning capability in llms via reinforcement learning},
  author={Guo, Daya and Yang, Dejian and Zhang, Haowei and Song, Junxiao and Wang, Peiyi and Zhu, Qihao and Xu, Runxin and Zhang, Ruoyu and Ma, Shirong and Bi, Xiao and others},
  journal={arXiv preprint arXiv:2501.12948},
  year={2025}
}

@article{wang2025timer1,
  title={Time-r1: Post-training large vision language model for temporal video grounding},
  author={Wang, Ye and Wang, Ziheng and Xu, Boshen and Du, Yang and Lin, Kejun and Xiao, Zihan and Yue, Zihao and Ju, Jianzhong and Zhang, Liang and Yang, Dingyi and others},
  journal={Advances in Neural Information Processing Systems},
  volume={38},
  pages={83330--83364},
  year={2026}
}

@misc{gemini38flash,
  title        = {{Gemini 3.8 Flash} Model Card},
  author       = {{Google DeepMind}},
  year         = {2026},
  howpublished = {\url{https://deepmind.google/models/model-cards/gemini-3-8-flash/}}
}

@misc{gemini31pro,
  title        = {{Gemini 3.1 Pro} Model Card},
  author       = {{Google DeepMind}},
  year         = {2026},
  howpublished = {\url{https://deepmind.google/models/model-cards/gemini-3-1-pro/}}
}

@misc{claudeopus5,
  title        = {{Claude Opus 5}},
  author       = {{Anthropic}},
  year         = {2026},
  howpublished = {\url{https://www.anthropic.com/claude}}
}

@misc{gpt56,
  title        = {Previewing {GPT-5.6 Sol}: A Next-Generation Model},
  author       = {{OpenAI}},
  year         = {2026},
  howpublished = {\url{https://openai.com/index/previewing-gpt-5-6-sol/}}
}

@article{bai2025qwen3vl,
  title={Qwen3-vl technical report},
  author={Bai, Shuai and Cai, Yuxuan and Chen, Ruizhe and Chen, Keqin and Chen, Xionghui and Cheng, Zesen and Deng, Lianghao and Ding, Wei and Gao, Chang and Ge, Chunjiang and others},
  journal={arXiv preprint arXiv:2511.21631},
  year={2025}
}

@misc{qwen35,
  title        = {{Qwen3.5}: Towards Native Multimodal Agents},
  author       = {{Qwen Team}},
  month        = feb,
  year         = {2026},
  howpublished = {\url{https://qwen.ai/blog?id=qwen3.5}}
}

@misc{qwen36,
  title        = {{Qwen3.6-27B}: Flagship-Level Coding in a {27B} Dense Model},
  author       = {{Qwen Team}},
  month        = apr,
  year         = {2026},
  howpublished = {\url{https://qwen.ai/blog?id=qwen3.6-27b}}
}

@misc{qwen38,
  title        = {{Qwen3.8-Max}: A New Bar for Coding and Cowork},
  author       = {{Qwen Team}},
  month        = aug,
  year         = {2026},
  howpublished = {\url{https://qwen.ai/blog?id=qwen3.8}}
}

@article{wang2025internvl35,
  title={Internvl3. 5: Advancing open-source multimodal models in versatility, reasoning, and efficiency},
  author={Wang, Weiyun and Gao, Zhangwei and Gu, Lixin and Pu, Hengjun and Cui, Long and Wei, Xingguang and Liu, Zhaoyang and Jing, Linglin and Ye, Shenglong and Shao, Jie and others},
  journal={arXiv preprint arXiv:2508.18265},
  year={2025}
}

@article{li2024llavaonevision,
  title={Llava-onevision: Easy visual task transfer},
  author={Li, Bo and Zhang, Yuanhan and Guo, Dong and Zhang, Renrui and Li, Feng and Zhang, Hao and Zhang, Kaichen and Zhang, Peiyuan and Li, Yanwei and Liu, Ziwei and others},
  journal={arXiv preprint arXiv:2408.03326},
  year={2024}
}

@article{zhang2024llavavideo,
  title={Llava-video: Video instruction tuning with synthetic data},
  author={Zhang, Yuanhan and Wu, Jinming and Li, Wei and Li, Bo and Ma, Zejun and Liu, Ziwei and Li, Chunyuan},
  journal={arXiv preprint arXiv:2410.02713},
  year={2024}
}

@article{zhang2025videollama3,
  title={Videollama 3: Frontier multimodal foundation models for image and video understanding},
  author={Zhang, Boqiang and Li, Kehan and Cheng, Zesen and Hu, Zhiqiang and Yuan, Yuqian and Chen, Guanzheng and Leng, Sicong and Jiang, Yuming and Zhang, Hang and Li, Xin and others},
  journal={arXiv preprint arXiv:2501.13106},
  year={2025}
}

@inproceedings{yu2025minicpmv45,
  title={Minicpm-v 4.5: Cooking efficient mllms via architecture, data, and training recipe},
  author={Yu, Tianyu and Wang, Zefan and Wang, Chongyi and Huang, Fuwei and Ma, Wenshuo and He, Zhihui and Cai, Tianchi and Chen, Weize and Huang, Yuxiang and Zhao, Ranchi and others},
  booktitle={Proceedings of the IEEE/CVF Conference on Computer Vision and Pattern Recognition},
  pages={11704--11715},
  year={2026}
}

@article{yang2025keyevl15,
  title={Kwai keye-vl 1.5 technical report},
  author={Yang, Biao and Wen, Bin and Ding, Boyang and Liu, Changyi and Chu, Chenglong and Song, Chengru and Rao, Chongling and Yi, Chuan and Li, Da and Zang, Dunju and others},
  journal={arXiv preprint arXiv:2509.01563},
  year={2025}
}

@inproceedings{clark2026molmo2,
  title={Molmo2: Open weights and data for vision-language models with video understanding and grounding},
  author={Clark, Christopher and Zhang, Jieyu and Ma, Zixian and Park, Jae Sung and Tripathi, Rohun and Lee, Sangho and Salehi, Mohammadreza and Ren, Jason and Kim, Chris Dongjoo and Yang, Yinuo and others},
  booktitle={Proceedings of the IEEE/CVF Conference on Computer Vision and Pattern Recognition},
  pages={28652--28668},
  year={2026}
}

@article{hong2025glm41v,
  title={Glm-4.5 v and glm-4.1 v-thinking: Towards versatile multimodal reasoning with scalable reinforcement learning},
  author={Hong, Wenyi and Yu, Wenmeng and Gu, Xiaotao and Wang, Guo and Gan, Guobing and Tang, Haomiao and Cheng, Jiale and Qi, Ji and Ji, Junhui and Pan, Lihang and others},
  journal={arXiv preprint arXiv:2507.01006},
  year={2025}
}

@article{azzolini2025cosmosreason1,
  title={Cosmos-reason1: From physical common sense to embodied reasoning},
  author={Azzolini, Alisson and Bai, Junjie and Brandon, Hannah and Cao, Jiaxin and Chattopadhyay, Prithvijit and Chen, Huayu and Chu, Jinju and Cui, Yin and Diamond, Jenna and Ding, Yifan and others},
  journal={arXiv preprint arXiv:2503.15558},
  year={2025}
}

@article{yu2025dapo,
  title={Dapo: An open-source llm reinforcement learning system at scale},
  author={Yu, Qiying and Zhang, Zheng and Zhu, Ruofei and Yuan, Yufeng and Zuo, Xiaochen and Yue, Yu and Dai, Weinan and Fan, Tiantian and Liu, Gaohong and Liu, Lingjun and others},
  journal={Advances in Neural Information Processing Systems},
  volume={38},
  pages={113222--113244},
  year={2026}
}

@article{liu2025drgrpo,
  title={Understanding r1-zero-like training: A critical perspective},
  author={Liu, Zichen and Chen, Changyu and Li, Wenjun and Qi, Penghui and Pang, Tianyu and Du, Chao and Lee, Wee Sun and Lin, Min},
  journal={arXiv preprint arXiv:2503.20783},
  year={2025}
}

@inproceedings{sheng2024hybridflow,
  title={Hybridflow: A flexible and efficient rlhf framework},
  author={Sheng, Guangming and Zhang, Chi and Ye, Zilingfeng and Wu, Xibin and Zhang, Wang and Zhang, Ru and Peng, Yanghua and Lin, Haibin and Wu, Chuan},
  booktitle={Proceedings of the Twentieth European Conference on Computer Systems},
  pages={1279--1297},
  year={2025}
}

@inproceedings{fu2024videomme,
  title={Video-mme: The first-ever comprehensive evaluation benchmark of multi-modal llms in video analysis},
  author={Fu, Chaoyou and Dai, Yuhan and Luo, Yongdong and Li, Lei and Ren, Shuhuai and Zhang, Renrui and Wang, Zihan and Zhou, Chenyu and Shen, Yunhang and Zhang, Mengdan and others},
  booktitle={2025 IEEE/CVF Conference on Computer Vision and Pattern Recognition (CVPR)},
  pages={24108--24118},
  year={2025},
  organization={IEEE}
}

@inproceedings{li2024mvbench,
  title={Mvbench: A comprehensive multi-modal video understanding benchmark},
  author={Li, Kunchang and Wang, Yali and He, Yinan and Li, Yizhuo and Wang, Yi and Liu, Yi and Wang, Zun and Xu, Jilan and Chen, Guo and Luo, Ping and others},
  booktitle={Proceedings of the IEEE/CVF conference on computer vision and pattern recognition},
  pages={22195--22206},
  year={2024}
}

@article{mangalam2023egoschema,
  title={Egoschema: A diagnostic benchmark for very long-form video language understanding},
  author={Mangalam, Karttikeya and Akshulakov, Raiymbek and Malik, Jitendra},
  journal={Advances in Neural Information Processing Systems},
  volume={36},
  pages={46212--46244},
  year={2023}
}

@inproceedings{xiao2021nextqa,
  title={Next-qa: Next phase of question-answering to explaining temporal actions},
  author={Xiao, Junbin and Shang, Xindi and Yao, Angela and Chua, Tat-Seng},
  booktitle={2021 IEEE/CVF Conference on Computer Vision and Pattern Recognition (CVPR)},
  pages={9772--9781},
  year={2021},
  organization={IEEE}
}

@inproceedings{buch2022revisiting,
  title={Revisiting the “video” in video-language understanding},
  author={Buch, Shyamal and Eyzaguirre, Crist{\'o}bal and Gaidon, Adrien and Wu, Jiajun and Fei-Fei, Li and Niebles, Juan Carlos},
  booktitle={2022 IEEE/CVF Conference on Computer Vision and Pattern Recognition (CVPR)},
  pages={2907--2917},
  year={2022},
  organization={IEEE}
}

@inproceedings{lei2022singleframe,
  title={Revealing single frame bias for video-and-language learning},
  author={Lei, Jie and Berg, Tamara and Bansal, Mohit},
  booktitle={Proceedings of the 61st Annual Meeting of the Association for Computational Linguistics (Volume 1: Long Papers)},
  pages={487--507},
  year={2023}
}

@article{cores2024tvbench,
  title={Lost in time: A new temporal benchmark for videollms},
  author={Cores, Daniel and Dorkenwald, Michael and Mucientes, Manuel and Snoek, Cees GM and Asano, Yuki M},
  journal={arXiv preprint arXiv:2410.07752},
  year={2024}
}

@inproceedings{liu2024tempcompass,
  title={Tempcompass: Do video llms really understand videos?},
  author={Liu, Yuanxin and Li, Shicheng and Liu, Yi and Wang, Yuxiang and Ren, Shuhuai and Li, Lei and Chen, Sishuo and Sun, Xu and Hou, Lu},
  booktitle={Findings of the Association for Computational Linguistics: ACL 2024},
  pages={8731--8772},
  year={2024}
}

@inproceedings{li2024vitatecs,
  title={Vitatecs: A diagnostic dataset for temporal concept understanding of video-language models},
  author={Li, Shicheng and Li, Lei and Liu, Yi and Ren, Shuhuai and Liu, Yuanxin and Gao, Rundong and Sun, Xu and Hou, Lu},
  booktitle={European Conference on Computer Vision},
  pages={331--348},
  year={2024},
  organization={Springer}
}

@article{zhang2024vinoground,
  title={Vinoground: Scrutinizing lmms over dense temporal reasoning with short videos},
  author={Zhang, Jianrui and Cai, Mu and Lee, Yong Jae},
  journal={arXiv preprint arXiv:2410.02763},
  year={2024}
}

@inproceedings{bagad2023testoftime,
  title={Test of time: Instilling video-language models with a sense of time},
  author={Bagad, Piyush and Tapaswi, Makarand and Snoek, Cees GM},
  booktitle={2023 IEEE/CVF Conference on Computer Vision and Pattern Recognition (CVPR)},
  pages={2503--2516},
  year={2023},
  organization={IEEE}
}

@article{patraucean2023perceptiontest,
  title={Perception test: A diagnostic benchmark for multimodal video models},
  author={Patraucean, Viorica and Smaira, Lucas and Gupta, Ankush and Recasens, Adria and Markeeva, Larisa and Banarse, Dylan and Koppula, Skanda and Malinowski, Mateusz and Yang, Yi and Doersch, Carl and others},
  journal={Advances in Neural Information Processing Systems},
  volume={36},
  pages={42748--42761},
  year={2023}
}

@article{xue2025arrowoftime,
  title={Seeing the arrow of time in large multimodal models},
  author={Xue, Zihui Sherry and Luo, Romy and Grauman, Kristen},
  journal={Advances in Neural Information Processing Systems},
  volume={38},
  pages={90925--90955},
  year={2026}
}

@article{feng2025videor1,
  title={Video-r1: Reinforcing video reasoning in mllms},
  author={Feng, Kaituo and Gong, Kaixiong and Li, Bohao and Guo, Zonghao and Wang, Yibing and Peng, Tianshuo and Wu, Junfei and Zhang, Xiaoying and Wang, Benyou and Yue, Xiangyu},
  journal={Advances in Neural Information Processing Systems},
  volume={38},
  pages={99114--99137},
  year={2026}
}

@article{cai2024temporalbench,
  title={Temporalbench: Benchmarking fine-grained temporal understanding for multimodal video models},
  author={Cai, Mu and Tan, Reuben and Zhang, Jianrui and Zou, Bocheng and Zhang, Kai and Yao, Feng and Zhu, Fangrui and Gu, Jing and Zhong, Yiwu and Shang, Yuzhang and others},
  journal={arXiv preprint arXiv:2410.10818},
  year={2024}
}

@inproceedings{shangguan2024tomato,
  title={Tomato: Assessing visual temporal reasoning capabilities in multimodal foundation models},
  author={Shangguan, Ziyao and Li, Chuhan and Ding, Yuxuan and Zheng, Yanan and Zhao, Yilun and Fitzgerald, Tesca and Cohan, Arman},
  booktitle={International Conference on Learning Representations},
  volume={2025},
  pages={7593--7734},
  year={2025}
}

@inproceedings{goyal2017something,
  title={The “something something” video database for learning and evaluating visual common sense},
  author={Goyal, Raghav and Kahou, Samira Ebrahimi and Michalski, Vincent and Materzynska, Joanna and Westphal, Susanne and Kim, Heuna and Haenel, Valentin and Fruend, Ingo and Yianilos, Peter and Mueller-Freitag, Moritz and others},
  booktitle={2017 IEEE international conference on computer vision (ICCV)},
  pages={5843--5851},
  year={2017},
  organization={IEEE}
}

@article{li2025videochatr1,
  title={Videochat-r1: Enhancing spatio-temporal perception via reinforcement fine-tuning},
  author={Li, Xinhao and Yan, Ziang and Meng, Desen and Dong, Lu and Zeng, Xiangyu and He, Yinan and Wang, Yali and Qiao, Yu and Wang, Yi and Wang, Limin},
  journal={arXiv preprint arXiv:2504.06958},
  year={2025}
}

@inproceedings{pickup2014arrow,
  title={Seeing the arrow of time},
  author={Pickup, Lyndsey C and Pan, Zheng and Wei, Donglai and Shih, YiChang and Zhang, Changshui and Zisserman, Andrew and Scholkopf, Bernhard and Freeman, William T},
  booktitle={Proceedings of the IEEE Conference on Computer Vision and Pattern Recognition},
  pages={2035--2042},
  year={2014}
}

@inproceedings{wei2018arrow,
  title={Learning and using the arrow of time},
  author={Wei, Donglai and Lim, Joseph J and Zisserman, Andrew and Freeman, William T},
  booktitle={Proceedings of the IEEE conference on computer vision and pattern recognition},
  pages={8052--8060},
  year={2018}
}

@article{chu2025sft,
  title={Sft memorizes, rl generalizes: A comparative study of foundation model post-training},
  author={Chu, Tianzhe and Zhai, Yuexiang and Yang, Jihan and Tong, Shengbang and Xie, Saining and Schuurmans, Dale and Le, Quoc V and Levine, Sergey and Ma, Yi},
  journal={arXiv preprint arXiv:2501.17161},
  year={2025}
}
}


\end{document}